\documentclass[letterpaper]{article} 
\usepackage[submission]{aaai25}  
\usepackage{times}  
\usepackage{helvet}  
\usepackage{courier}  
\usepackage[hyphens]{url}  
\usepackage{graphicx} 
\usepackage{natbib}  
\usepackage{caption} 
\usepackage{algorithm}
\usepackage{algorithmic}

\usepackage{amsmath} 
\usepackage{amssymb}
\usepackage{stmaryrd}
\usepackage{makecell}
\usepackage{booktabs}
\usepackage{multirow}
\usepackage{cleveref}

\crefname{figure}{Figure}{Figures}
\crefname{table}{Table}{Tables}
\crefname{section}{Section}{Sections}
\crefname{equation}{Equation}{Equations}

\newcommand{\En}[1]{\llbracket #1 \rrbracket}

\DeclareMathOperator{\rrot}{rot}
\newcommand{\rot}[1]{\rrot(#1)}

\newcommand{\ColTile}{$\mathtt{ColTile}$}
\newcommand{\PolyComp}{$\mathtt{PolyComp}$}

\usepackage{newfloat}
\usepackage{listings}
\DeclareCaptionStyle{ruled}{labelfont=normalfont,labelsep=colon,strut=off} 
\floatstyle{ruled}
\newfloat{listing}{tb}{lst}{}
\floatname{listing}{Listing}
\title{Efficient Privacy-Preserving KAN Inference Using Homomorphic Encryption}
\author {
    Zhizheng Lai,
    Yufei Zhou,
    Peijia Zheng\thanks{Corresponding author}
    Lin Chen
}
\affiliations {
    Guangdong Provincial Key Laboratory of Information Security Technology, \\School of Computer Science and Engineering, \\Sun Yat-sen University\\
    laizhzh6@mail2.sysu.edu.cn, zhouyf55@mail2.sysu.edu.cn, zhpj@mail.sysu.edu.cn, chenlin69@mail.sysu.edu.cn
}

\usepackage{bibentry}

\usepackage{color,xcolor}

\begin{document}
\maketitle

\begin{abstract}

The recently proposed Kolmogorov-Arnold Networks (KANs) offer enhanced interpretability and greater model expressiveness. However, KANs also present challenges related to privacy leakage during inference. Homomorphic encryption (HE) facilitates privacy-preserving inference for deep learning models, enabling resource-limited users to benefit from deep learning services while ensuring data security. Yet, the complex structure of KANs, incorporating nonlinear elements like the SiLU activation function and B-spline functions, renders existing privacy-preserving inference techniques inadequate. To address this issue, we propose an accurate and efficient privacy-preserving inference scheme tailored for KANs. Our approach introduces a task-specific polynomial approximation for the SiLU activation function, dynamically adjusting the approximation range to ensure high accuracy on real-world datasets. Additionally, we develop an efficient method for computing B-spline functions within the HE domain, leveraging techniques such as repeat packing, lazy combination, and comparison functions. We evaluate the effectiveness of our privacy-preserving KAN inference scheme on both symbolic formula evaluation and image classification. The experimental results show that our model achieves accuracy comparable to plaintext KANs across various datasets and outperforms plaintext MLPs. Additionally, on the CIFAR-10 dataset, our inference latency achieves over $7\times$ speedup compared to the naive method.

\end{abstract}

%

\section{Introduction}


In recent years, deep learning has made notable advancements. However, training a model requires substantial data and computational power, which is often impossible for resource-limited companies and individuals. Additionally, model owners are frequently hesitant to share their deep models due to potential breaches of intellectual property or privacy concerns \cite{jegorova2022survey}.
Using Machine Learning as a Service (MLaaS) offers a potential solution, allowing users to leverage encrypted deep learning services. However, user data, such as images or bills, often contain personal information, which raises significant concerns about privacy exposure in MLaaS. 

To address these security issues, deep private inference has garnered considerable attention in recent years. By utilizing Secure Multiparty Computation (MPC) \cite{patra2021aby2} and Homomorphic Encryption (HE) \cite{marcolla2022survey}, data owners and model owners can perform inference tasks without compromising each other's privacy \cite{knott2021crypten,lee2022low}.
MPC-based privacy-preserving inference schemes often require substantial communication overhead \cite{hao2022iron,pang2024bolt}. In contrast, in HE-based privacy-preserving inference frameworks \cite{lee2022low,ran2023spencnn,zimerman2024converting}, there is no need for interaction between the user and the server. Specifically, the user encrypts the input and sends it to the server, which then executes the inference algorithm and returns the encrypted result to the user. In this paper, we focus on HE-based solutions.

\begin{figure}[!t]
\centering
\includegraphics[width=0.95\columnwidth]{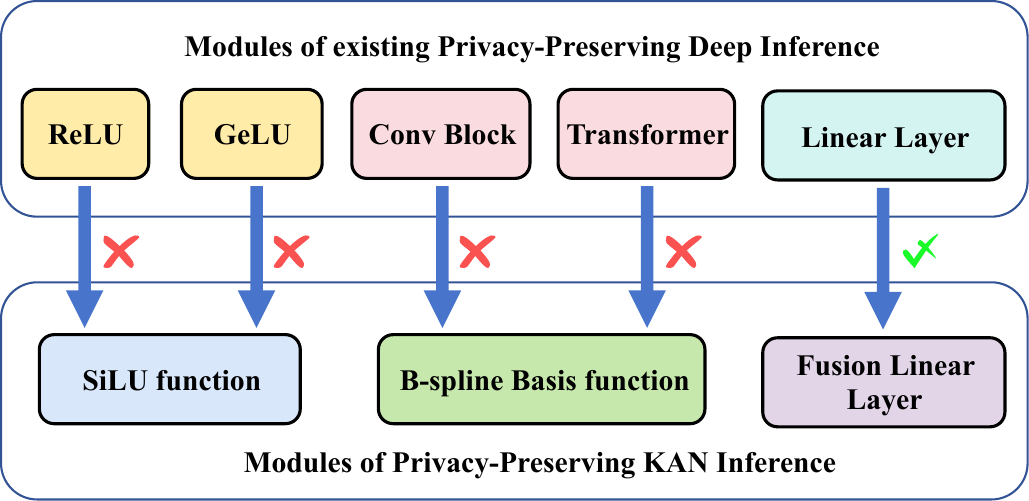} 
\caption{Obstacles in transitioning from privacy-preserving inference of existing neural networks to privacy-preserving inference of KAN.}
\label{figInddd}
\end{figure}

Although many deep private inference schemes have been proposed recently, the interpretability of deep models remains an open problem. This year, Kolmogorov-Arnold Networks (KANs) were introduced with the aim of offering greater interpretability. KANs incorporate additional non-linear elements, which increase computational demands during training.
As shown in \Cref{figInddd}, due to structural differences between KAN networks and traditional neural network architectures, \textit{existing HE-based privacy-preserving inference methods of deep learing cannot be directly applied to KANs}. First, KAN inference requires the computation of various activation functions, like SiLU. Additionally, KAN inference necessitates the approximation of B-splines, for which no HE-based computation currently exists \cite{knott1999interpolating}. HE schemes capable of performing Boolean operations, such as TFHE \cite{chillotti2020tfhe}, can compute arbitrary functions but are highly inefficient. As a result, most privacy-preserving inference methods \cite{lee2022low,ran2023spencnn} rely on arithmetic fully homomorphic encryption (FHE) schemes, such as RNS-CKKS \cite{cheon2019full}, to implement privacy-preserving KAN inference. These schemes support only homomorphic addition and multiplication, necessitating polynomial approximations of activation functions and B-splines.

While ReLU activation functions have been well-approximated in current privacy-preserving inference work \cite{lee2022low,ran2023spencnn} through high-precision sign function approximations \cite{lee2021minimax}, approximating the SiLU function is more challenging due to its lack of a direct relationship with the sign function. Although better polynomial approximations can be achieved through training \cite{ao2023autofhe}, the training cost for KAN networks is higher than for other neural network models due to the additional activation functions. 
Furthermore, in many cases, the server only has access to the model, not the original training data.

To address these challenges, we propose a novel task-specific activation function approximation method that dynamically adjusts the approximation range to suit different KAN networks. This method leverages Chebyshev's inequality, focusing on regions with dense data distribution while avoiding emphasis on sparse edge regions. Comparisons with existing approximation methods, such as Remez~\cite{zimerman2024converting}, show superior model performance. Additionally, we develop an approximation method for B-spline basis functions using comparison functions based on minimax composite polynomials~\cite{lee2021minimax}. We introduce a repeat packing technique to parallelize the computation of B-spline basis functions, enhancing computational efficiency. Moreover, we employ lazy combination techniques to eliminate extra time costs associated with rearranging encrypted data, achieving a $5.54\times$ speedup compared to the naive method under larger parameter settings.

\begin{itemize}
    \item We propose a novel method for approximating activation functions by dynamically determining the approximation range and using weighted least squares. This approach yields a polynomial approximation of the SiLU activation function that performs effectively on real datasets.
    \item We introduce an efficient approach for computing B-spline functions in the HE domain. By utilizing techniques such as repeat packing, lazy combination, and comparison functions, we achieve highly accurate B-spline function computation.
    \item Our experiments validate that our HE KAN maintains competitive performance in both symbolic formula evaluation and image classification. Additionally, on the CIFAR-10 dataset, our inference latency achieves over $7\times$ speedup compared to the naive method.
\end{itemize}

\section{Related Work}

\subsection{KANs in the Plaintext Domain}
KAN, as a promising alternative to MLP, has been integrated with popular models by various researchers. In~\cite{zhang2024graphkanenhancingfeatureextraction}, Zhang et al. combined KAN with Graph Neural Networks, utilizing KAN for feature extraction and experimentally demonstrating the effectiveness of their proposed approach. Xu et al.~\cite{xu2024fourierkangcffourierkolmogorovarnoldnetwork} introduced KAN to enhance the message-passing process in Graph Collaborative Filtering, proposing a more efficient recommendation model. In~\cite{genet2024temporalkolmogorovarnoldtransformertime}, Genet et al. incorporated Temporal KAN into the Temporal Fusion Transformer to further improve the model's ability to handle time series data. Li et al.~\cite{li2024ukanmakesstrongbackbone} combined KAN with U-Net for the segmentation and generation of medical images.
However, these works primarily focus on the performance of KAN within neural networks and do not address the issues of security and privacy leakage in practical applications.

\subsection{Privacy-Preserving Deep Inference}
In 2016, Ran et al. introduced CryptoNets~\cite{gilad2016cryptonets}, achieving privacy-preserving inference based on HE, though with very low efficiency. Since then, researchers have continued to improve inference performance on convolutional networks~\cite{chou2018faster,brutzkus2019low,benaissa2021tenseal}, though the network structures they implemented are relatively simple. Both Lee et al.\cite{lee2022low} and Ran et al.\cite{ran2023spencnn} employ more efficient ciphertext packing methods to achieve privacy-preserving inference for complex convolutional networks. Zimerman et al.~\cite{zimerman2024converting} achieved privacy-preserving inference for Transformers. Additionally, researchers have combined Graph Convolutional Networks (GCN) with HE to enhance privacy protection for cloud-based graph data inference~\cite{ran2022cryptogcn,peng2024lingcn,ran2024penguin}.
There are also many privacy-preserving inference schemes based on MPC~\cite{juvekar2018gazelle,srinivasan2019delphi,hao2022iron,pang2024bolt}, but the communication overhead of MPC-based schemes is enormous, making them unsuitable for resource-constrained users.
KANs differs significantly from current deep models. The aforementioned methods rely heavily on specific model structures, making them unsuitable for direct application to KAN privacy-preserving inference. Therefore, it is necessary to design a new method to achieve privacy-preserving inference for KAN.

\section{Preliminaries}
\subsection{Kolmogorov-Arnold Networks}


Unlike traditional neural networks that use fixed activation functions, KANs employs learnable activation functions at the network's edges, enabling each weight parameter to be replaced by a univariate function. However, identifying suitable basis functions can be challenging, as many cannot be represented effectively. To address this, KANs utilizes spline functions for parametric approximation, providing significant flexibility. This method allows for modeling complex functions with fewer parameters, thereby enhancing the model's interpretability.

The flexibility of spline functions enables them to adaptively model complex relationships in data by adjusting their shape, minimizing approximation errors, and improving the network's ability to learn subtle patterns from high-dimensional datasets.

In most cases, $\operatorname{spline}(x)$ is parametrized as a linear combination of B-splines as follows~\cite{liu2024kan}:
\begin{equation}\label{eq:splinex}
    \operatorname{spline}(x)=\sum_m^n c_m B_{m,k}(x)
\end{equation}
where $c_m$ are trainable parameters, and $B_{m,k}(x)$ are the $k$-order B-spline basis functions at the knots. The $k$-order B-spline basis functions are defined recursively as follows:
\begin{equation}
\label{eq:comForSpline}
    B_{m,0}(x)=\begin{cases}1&\text{if }t_m\leq x<t_{m+1}\\0&\text{otherwise}\end{cases}
\end{equation}
\begin{equation}
\label{eq:bspline}
\begin{split}
    B_{m,k}(x) & = \frac{x-t_m}{t_{m+k}-t_m}B_{m,k-1}(x) \\ & +\frac{t_{m+k+1}-x}{t_{m+k+1}-t_{m+1}}B_{m+1,k-1}(x)
\end{split}
\end{equation}
where $t_1,\cdots,t_{n+k}$ form a non-decreasing sequence of knots, which determine the domain and influence of the basis functions.

The overall network in KAN can be expressed by the following equations:
\begin{equation}\label{kaneq1}
    \phi(x)=w_bb(x)+w_s\operatorname{spline}(x)
\end{equation}

\begin{equation}\label{kaneq2}
    b(x)=\operatorname{silu}(x)= \frac{x}{1+e^{-x}}
\end{equation}

    

\subsection{Homomorphic Encryption}


Following current work in privacy-preserving inference~\cite{lee2022low,ran2023spencnn}, we use RNS-CKKS~\cite{cheon2019full} as the underlying cryptographic scheme. Below, we briefly introduce the main operations in HE.

RNS-CKKS encrypts a vector at a time using Single Instruction, Multiple Data (SIMD). To clearly describe the operations of HE, we use bold lowercase letters to represent vectors, and $\En{\cdot}$ to denote encrypted ciphertexts. For example, we use $\mathbf{x}$ to represent the vector $(x_1,x_2,\cdots,x_n)$, and $\En{\mathbf{x}}$ to denote its ciphertext.
We use $\oplus$, $\ominus$, $\otimes$, and $\rot$ represent homomorphic ciphertext addition, subtraction, multiplication, and rotation.

\begin{align}
\En{\mathbf{x}} \oplus  \En{\mathbf{y}}  &= \En{\left(x_1 + y_1,x_2 + y_2,\cdots,x_{n} + y_{n}\right)}\\
\En{\mathbf{x}} \ominus  \En{\mathbf{y}}  &= \En{\left(x_1 - y_1,x_2 - y_2,\cdots,x_{n} - y_{n}\right)}\\
\En{\mathbf{x}} \otimes  \En{\mathbf{y}}  &= \En{(x_1 \times y_1,x_2 \times y_2,\cdots,x_{n} \times y_{n})} \\
\En{\mathbf{x}} \otimes \mathbf{y}  &= \En{\left(x_1 \times y_1,x_2 \times y_2,\cdots,x_{n} \times y_{n}\right)}\\
\rot{\mathbf{x},t} &= \En{\left(x_{t+1},\cdots,x_{k},x_1,\cdots,x_{t}\right)}, t \in \mathbb{Z}
\end{align}
where $t$ is the rotation step. When $t > 0$, the ciphertext rotates left; when $t < 0$, it rotates right. A more detailed theoretical analysis and security assessment can be found in~\cite{cheon2017homomorphic,cheon2019full}.








\subsection{Threat Model}
The threat model in this paper is similar to that of previous HE-based privacy-preserving inference schemes~\cite{lee2022low,ran2023spencnn}. We assume a cloud-based machine learning service scenario where the cloud server hosts a pre-trained KAN model with plaintext weights. Clients send their sensitive data to the cloud server to obtain inference results from the network model. We assume the cloud server is honest-but-curious, meaning it follows the protocol but may attempt to extract users' private information through additional computations. To protect the privacy of the client's sensitive data, the client can encrypt the data using HE before sending it to the cloud server for privacy-preserving inference. The cloud server performs computations on the ciphertext without decrypting it, and the client decrypts the inference results using their private key. This process ensures that no information about the input data is revealed to the cloud.


\section{Method}
\begin{figure}[t]
    \centering
    \includegraphics[width=0.9\columnwidth]{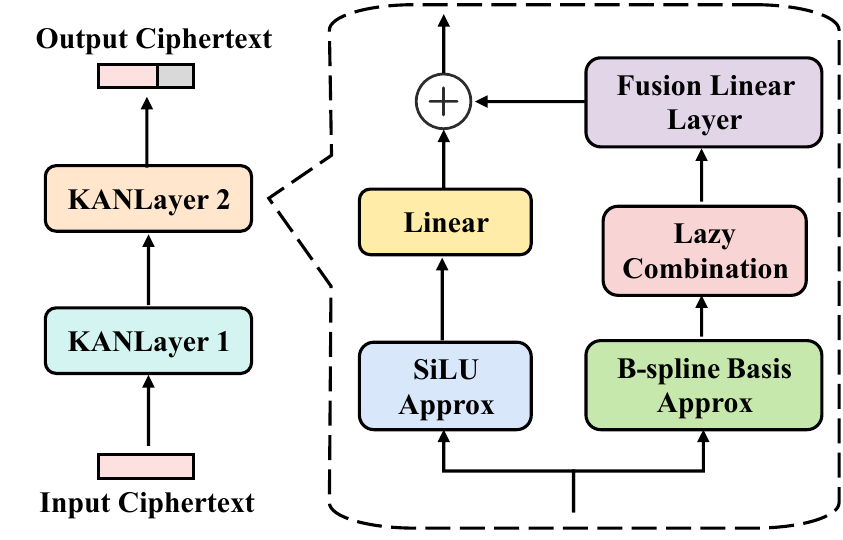}
    \caption{
        Structure of HE-friendly KAN. The specific structure of the KANLayer is shown within the dashed box.
    }
    \label{KANLayer}
\end{figure}

Similar to MLPs, the KANs is composed of multiple KANLayers stacked linearly, with each KANLayer having an input dimension of $n_{i}$ and an output dimension of $n_{o}$. Therefore, by approximating a single KANLayer, we can approximate the entire network. The overview of our HE-friendly KAN is shown in \cref{KANLayer}.

In our HE KAN, the input is a three-dimensional real-valued tensor $\mathbf{A} \in \mathbb{R}^{h\times w\times c}$, where $h$ and $w$ represent the height and width of the input, respectively, and $c$ represents the number of channels. We use a raster scan order~\cite{lee2022low} to convert the tensor $\mathbf{A}$ into a vector $\mathbf{x}$ of length $n_{i}=h\cdot w\cdot c$, which is encrypted using CKKS-RNS as $\En{\mathbf{x}}$. Each KANLayer includes SiLU, B-spline, and Linear Layer. Below, we describe the approximation method for each component in detail.

\subsection{SiLU Approximation}
Previous approximations of the ReLU function utilized the sign function, which allows the input to be scaled to the $[-1,1]$ range, resulting in a good approximation. However, SiLU cannot be approximated using sign or comparison functions. The biggest challenge in approximating SiLU is determining the appropriate approximation range.

The input range of the activation function varies across different datasets~\cite{lee2023hetal}. We estimate the mean $\mu$ and variance $\sigma^2$ of the activation function's input range by feeding the dataset into a pre-trained network. Then, we use Chebyshev's inequality to estimate the approximate distribution of the activation function's input values.

According to Chebyshev's inequality,
\begin{equation}
\label{eq:chebyshev}
    P\left\{\left|X-\mu\right|\geqslant\varepsilon\right\}\leqslant\frac{\sigma^{2}}{\varepsilon^{2}}
\end{equation}
when we set $\varepsilon=5\sigma$, we assume that approximately $96\%$ of the data falls within the $[(\mu-5\sigma, \mu+5\sigma)]$ interval. Therefore, we use $[\max(\mu-5\sigma, x_{\text{min}}), \min(\mu+5\sigma, x_{\text{max}})]$ as the approximation range to ensure that most data lies within this interval.

Next, we uniformly sample points within the approximation range and use weighted least squares to fit the activation function. We focus on fitting the regions with dense data distribution and de-emphasize the sparse edge regions. This approach allows us to obtain a polynomial that better approximates the original activation function under the same polynomial degree constraints.
Let the input be $x_i$, the output of the activation function be $y_i$, and the output of the polynomial be $f(x_i)$. The optimization objective of weighted least squares is:
\begin{equation}
    \min(\sum_{i=1}^nw_i(y_i-f(x_i))^2)
\end{equation}
where $n$ is the number of sampled points, and $w_i$ is the weight. The closer the data is to the mean, the higher the weight we assign. For example, the sum of squared residuals of data points within the $[\mu-3\sigma, \mu+3\sigma]$ interval is given a weight $w_i$ of 10, while those outside this interval are given a weight of 1. We then solve for the polynomial coefficients to obtain the approximate polynomial for the activation function.

\subsection{B-spline Basis Approximation}
\label{sec:bsplineBasis}

We observed that the calculation of B-spline basis functions involves many parallelizable components. Therefore, we use Repeat Packing to fully leverage the SIMD capabilities of HE to accelerate B-spline calculations.

Let the input be a vector $\mathbf{x}$ of length $n_i$. After repeat packing, it becomes $\En{\mathbf{x}|\mathbf{x}|\mathbf{x}\dots|\mathbf{x}}$, denoted as $\En{\mathbf{x}}_{\text{rep}}$, where $\mathbf{x}$ is repeated $g+2k$ times, with $g$ being the number of B-spline grids and $k$ being the B-spline degrees, satisfying $n_{i}\cdot (g+2k)\leq \frac{N}{2}$.

Repeat packing data in ciphertexts is not easy, as ciphertext rotations are computationally expensive. We designed a fast packing method that reduces the number of rotations needed for repeat packing from $g+2k$ to $\lceil\log_2(g+2k)\rceil$. We first rotate $\En{\mathbf{x}}$ to the right by $n_i$ steps, then add it to the original $\En{\mathbf{x}}$ to obtain a temporary ciphertext $\En{\mathbf{x}|\mathbf{x}}$. The above steps are repeated on the temporary ciphertext, with rotation steps of $n_i, 2n_i, 4n_i,\cdots$. The specific algorithm is shown in \Cref{alg:repeatpacking}.

\begin{algorithm}[tb]
\caption{Fast Repeat Packing} 
\label{alg:repeatpacking}
\textbf{Input}: ciphertext $\En{\mathbf{x}}$,  grid parameters $g$ and $k$, input size $n_i$\\
\textbf{Output}: packed ciphertext $\En{\mathbf{x}}_{\text{rep}}$
\begin{algorithmic}[1] 

\STATE $\mathbf{m}=(1,1,\cdots,0,\dots)$ \COMMENT{There are $n_i$ ones in $\mathbf{m}$.}

\STATE $\En{\mathbf{x}}_{\text{rep}} \leftarrow \En{\mathbf{x}} \otimes \mathbf{m}$ 

\FOR{$j \leftarrow 1$ to $\lceil\log_2(g+2\cdot k)\rceil$}  
    \STATE $\En{\mathbf{x}}_{\text{rep}} \leftarrow \rot{\En{\mathbf{x}_{\text{rep}}}, -n_i\cdot 2^{j-1}} \oplus \En{\mathbf{x}}_{\text{rep}}$
\ENDFOR

\STATE \textbf{return} $\En{\mathbf{x}}_{\text{rep}}$ 
\end{algorithmic}
\end{algorithm}

\begin{figure}[t]
    \centering
    \includegraphics[width=0.9\columnwidth]{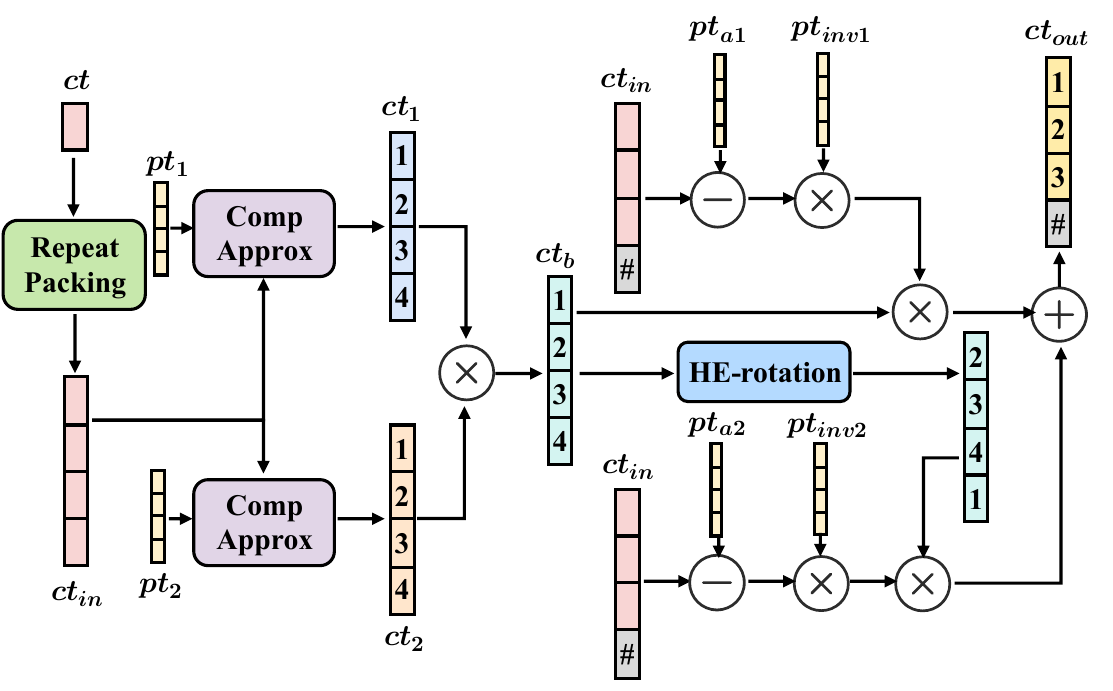}
    \caption{
        Toy example of B-spline Basis Approximation when $g=2$ and $k=1$.
    }
    \label{B-splineBases}
\end{figure}

To simplify the description, we define the following operations:
\begin{itemize}
    \item \ColTile$(\mathbf{A}, l, r)$: Concatenates columns $l$ to $r-1$ of matrix $\mathbf{A}$ into a vector.
    \item \PolyComp$(a, b)$: Returns $1$ if $a>b$, $0$ if $a<b$, and $\frac{1}{2}$ if $a=b$. We implement this using the method from~\cite{lee2021minimax}.
\end{itemize}

Recall \cref{eq:comForSpline} and \cref{eq:bspline}. We first need to determine whether the input $\mathbf{x}$ falls within the interval $[t_m,t_{m+1}]$, and then recursively compute the basis functions.

We combine all the intervals into a weight matrix $\mathbf{G}$. Then, using \ColTile, we obtain a vector $\mathbf{g_1}$ containing the left endpoints of all intervals $[t_m,t_{m+1}]$ and a vector $\mathbf{g_2}$ containing the right endpoints. Next, we use \PolyComp\ to compare the packed input $\En{\mathbf{x}}_{\text{rep}}$ with $\mathbf{g_1}$ and $\mathbf{g_2}$, and then recursively calculate the basis functions according to \cref{eq:bspline}.

It should be noted that \PolyComp\ operates within the range $[-1,1]$. If the input exceeds this range, the approximation may become inaccurate. Therefore, we scale the input to the $[-1,1]$ range. We estimate the approximate range of the input using the training data, and represent the input within a larger range $[-R,R]$ to avoid overflow. When using \PolyComp\ to compute intervals, there may be cases where $x$ is exactly equal to the interval endpoint, such as $x=t_m$. In this case, \PolyComp\ returns $\frac{1}{2}$, which does not match \cref{eq:comForSpline}. We can achieve the correct result by comparing $2x$ with $2t_m+1$. However, the probability of equality is very small, so its impact on the overall approximation error is negligible. Therefore, in practice, we can ignore the equality case, similar to how existing papers approximate ReLU~\cite{lee2022low}. The specific algorithm is shown in \Cref{alg:bSplineBasis}.

\begin{algorithm}[t]
\caption{B-spline Basis Approximation} 
\label{alg:bSplineBasis}
\textbf{Input}: packed ciphertext $\En{\mathbf{x}}_{\text{rep}}$, grid parameters $g$ and $k$, input size $n_i$, weight matrix $\mathbf{G}$, bound $R$\\
\textbf{Output}: B-spline basis result $\En{\mathbf{b}}$
\begin{algorithmic}[1] 
\STATE $r \leftarrow g + 2k + 1$
\STATE $\mathbf{g_1} \leftarrow $\ColTile$(\mathbf{G}, 1, r)$
\STATE $\mathbf{g_2} \leftarrow $\ColTile$(\mathbf{G}, 2, r+1)$
\STATE $\En{\mathbf{x_1}} \leftarrow $\PolyComp$((\En{\mathbf{x}}_{\text{rep}} \ominus \mathbf{g_1}) \otimes \frac{1}{2R},0)$

\STATE $\En{\mathbf{x_2}} \leftarrow $\PolyComp$((\mathbf{g_2} \ominus \En{\mathbf{x}}_{\text{rep}}) \otimes \frac{1}{2R},0)$
\STATE $\En{\mathbf{b}} \leftarrow \En{\mathbf{x_1}} \otimes \En{\mathbf{x_2}}$
\FOR{$j \leftarrow 1$ to $k$}
    \STATE $\mathbf{t_1} \leftarrow $\ColTile$(\mathbf{G}, 1, r-j))$
    \STATE $\mathbf{t_2} \leftarrow $\ColTile$(\mathbf{G}, j+1, r)$
    
    \STATE $\En{\mathbf{b_1}} \leftarrow  \frac{1}{\mathbf{t_2} - \mathbf{t_1}} \otimes (\En{\mathbf{x}}_{\text{rep}} \ominus \mathbf{t_1}) \otimes \En{\mathbf{b}} $
    \STATE $\mathbf{t_3} \leftarrow $\ColTile$(\mathbf{G}, j+2, r+1)$
    \STATE $\mathbf{t_4} \leftarrow $\ColTile$(\mathbf{G}, 2, r-j+1)$
    \STATE $\En{\mathbf{b_2}} \leftarrow \frac{1}{\mathbf{t_3} - \mathbf{t_4}} \otimes (\mathbf{t_3} \ominus \En{\mathbf{x}}_{\text{rep}}) \otimes \rot{\En{\mathbf{b}}, n_i} $
    \STATE $\En{\mathbf{b}} \leftarrow \En{\mathbf{b_1}} \oplus \En{\mathbf{b_2}}$
\ENDFOR
\STATE \textbf{return} $\En{\mathbf{b}}$
\end{algorithmic}
\end{algorithm}

\subsection{Lazy Combination and Fusion Linear Layer}

In the previous section, we described how to approximate B-spline basis functions. However, according to \cref{eq:splinex}, we need to multiply each basis function by a coefficient $c_m$ and sum them to obtain the final B-spline function approximation. For a given input $x$, this is an inner product between the coefficient vector $\mathbf{c}$ and the basis function vector $\mathbf{b}$. Since we are using SIMD, computing the inner product requires rotations, which are expensive in the HE domain. Therefore, we adopt a lazy combination approach, postponing the computation of $\operatorname{spline(x)}$ until later to reduce the number of rotations. Specifically, we combine $\mathbf{c}$ with the weight coefficients $\mathbf{W_s}$ of the subsequent linear layer to obtain a new set of weights. This reduces the number of multiplications and rotations, improving computational efficiency.

The weight matrix $\mathbf{W_s}$ has dimensions $n_o \times n_i$. When the input size is $n_i$, the coefficients of the basis functions form a matrix $\mathbf{C}$ of size $n_i \times (g+k)$. After combining $\mathbf{W_s}$ and $\mathbf{C}$, we obtain a large matrix $\mathbf{W^\prime}$ of size $n_o \times n_i \cdot(g+k)$. Therefore, following the method of~\cite{efficient-kan}, we directly use $\mathbf{W^\prime}$ during inference instead of the linear layer weights $\mathbf{W_s}$ and $\mathbf{C}$.

However, due to the repeat encoding introduced during B-spline basis function computation, the result is in \ColTile\ format, with an output vector of length $n_i\cdot(g+k)$. We need to repack the result of the basis function approximation into row-major order to take advantage of SIMD for fast matrix operations. This is essentially a ciphertext reordering problem, requiring multiplication by a permutation matrix $\mathbf{P}$ of size $n_i\cdot (g+k) \times n_i\cdot (g+k)$. The method for generating the permutation matrix is provided in \Cref{alg:permuta}.

To reduce multiplicative depth and the number of rotations, lazy combination is also used as mentioned above. We multiply the permutation matrix $\mathbf{P}$ with the weight coefficients $\mathbf{W^\prime}$. The final fusion linear layer weight is then given by:
\begin{equation}
    \mathbf{W_{f}}=\mathbf{W^\prime} \times \mathbf{P}
\end{equation}

The computation of the fusion linear layer, as well as the linear layer following SiLU, involves matrix multiplication. We optimize this using the baby-step giant-step (BSGS) strategy~\cite{halevi2021bootstrapping}, reducing the number of ciphertext rotations required for matrix multiplication and lowering the time cost of the linear layer.


\begin{algorithm}[t]
\caption{Generate Permutation Matrix} 
\label{alg:permuta}
\textbf{Input}: number of rows $n_r$, number of columns $n_c$ \\
\textbf{Output}: permutation matrix $\mathbf{P}$
\begin{algorithmic}[1] 
\STATE Initialize $\mathbf{P}$ using zeros.\COMMENT{The shape is  $n_r  n_c \times n_r n_c$}
\FOR{$r \leftarrow 1$ to $n_r$}
    \FOR{$c \leftarrow 1$ to $n_c$}
        \STATE $j_c \leftarrow (c - 1) \times n_r + r$
        \STATE $j_r \leftarrow (r - 1) \times n_c + c$
        \STATE $P[j_r][j_c] \leftarrow 1$
    \ENDFOR
\ENDFOR
\STATE \textbf{return} $P$
\end{algorithmic}
\end{algorithm}



\section{Experimental Evaluation}

\subsection{Experiment Setup}

\subsubsection{Environment.}
We conducted our experiments on a machine equipped with an Intel Xeon Gold 6145 CPU @ 2.00GHz and 512 GB RAM. And we used C++ and the Microsoft SEAL version 3.6.6~\cite{sealcrypto} library to implement the privacy-preserving KAN inference based on the RNS-CKKS scheme \cite{cheon2019full}.

\subsubsection{Encryption Parameters.}
We set the degree of the polynomial modulus to $2^{16}$ and used parameters recommended for a 128-bit security level according to \cite{albrecht2021homomorphic}. The bit-lengths of the base modulus and special modulus were set to $51$, while the bit-length of the scale factor was set to $46$. The multiplicative depth was set to $20$. For the Bootstrapping parameters, we used the same settings as in~\cite{lee2022low}.

\subsubsection{Datasets and Metric.}
We represent the KAN network structure as an array $(n_1, n_2, n_3)$, denoting the number of nodes per layer.
The datasets used in the experiments were MNIST~\cite{lecun1998gradient}, Fashion-MNIST~\cite{xiao2017fashion}, and CIFAR-10~\cite{krizhevsky2009learning}, and  we used KAN models with $(728, 64, 10)$, $(728, 128, 10)$, and $(3072, 128, 10)$, respectively. Note that when evaluating the model's inference performance, we used the models trained in the plaintext domain. So we just used  inference accuracy to evaluate model performance. For the comparison of effectiveness of approximation and symbolic formulas, we used RMSE as in \cite{liu2024kan}. To minimize errors, our experimental metrics were obtained by averaging the results of 10 trials.

\subsection{SiLU Approximation}

\subsubsection{Approximation Interval and Polynomial Degree.}

\begin{table}
\centering
\begin{tabular}{c|c|cc}
    \toprule
    Dataset & Degree & Acc. $\uparrow$ & RMSE $\downarrow$ \\
    \midrule
    \multirow{4}{*}{MNIST}& 8 & 96.82 & 0.307  \\
    & 9 & 95.87 & 0.183  \\
    & 10 & \textbf{97.17} & 0.166  \\
    & 11 & 96.40 & \textbf{0.153}  \\
    \midrule
    \multirow{4}{*}{FMNIST} & 8 & 89.26 & 0.158  \\
    & 9 & 89.32 & 0.110  \\
    & 10 & \textbf{89.35} & 0.093  \\
    & 11 & 89.29 & \textbf{0.085}  \\
    \midrule
    \multirow{4}{*}{CIFAR-10} & 13 & 57.08 & 0.053 \\
    & 14 & 57.21 & 0.046 \\
    & 15 & \textbf{57.38} & 0.042 \\
    & 16 & 57.03 & \textbf{0.025} \\
    \bottomrule
\end{tabular}
\caption{Comparison of model performance for SiLU approximation with different polynomial degrees across various datasets using our approximation method.}
\label{Tabdegree}
\end{table}

Since the input range of the activation function varies across different datasets~\cite{lee2023hetal}, different tasks require different approximation ranges. According to \cref{eq:chebyshev}, we estimated the optimal approximation interval using the $5\sigma$. We used approximation ranges of $[-12.4, 14.74]$ for MNIST, $[-9.77, 11.01]$ for Fashion-MNIST, and $[-11.90, 10.99]$ for CIFAR-10. 
Choosing the appropriate polynomial degree for approximation is also crucial for different tasks. \cref{Tabdegree} shows the impact of different polynomial degrees on accuracy across datasets. For better model performance, we select a polynomial degree of $10$ for the MNIST and Fashion-MNIST datasets, and a degree of $15$ for the CIFAR-10 dataset.

\subsubsection{Comparison of Different Approximation Methods.}

\begin{figure}[t]
\centering
\includegraphics[width=0.9\columnwidth]{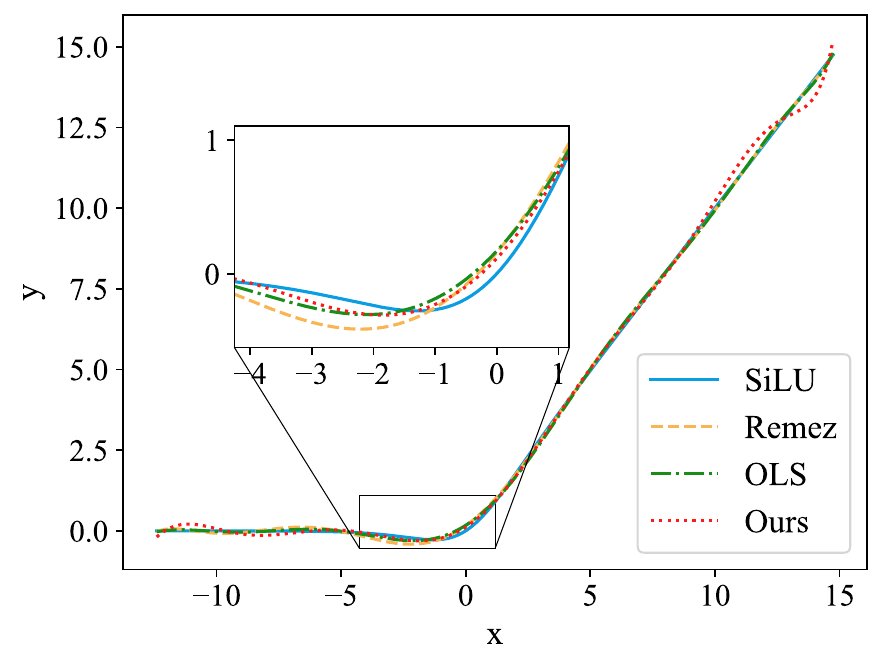} 
\caption{Comparison of our method with Remez \cite{zimerman2024converting} and OLS \cite{zheng2022keyword} for fitting the SiLU activation function on the MNIST dataset.}
\label{figSilu}
\end{figure}

\begin{table}[t]
\centering
\begin{tabular}{l|l|ccc}
    \toprule
    Method & Metric & MNIST & FMNIST & CIFAR-10 \\
    \midrule
    \multirow{2}{*}{Remez} & RMSE $\downarrow$ & 0.087 & 0.046 & 0.016 \\
    & Acc. $\uparrow$ & 96.76 & 89.16 & 57.17 \\
    \midrule
    \multirow{2}{*}{OLS} & RMSE $\downarrow$ & 0.071 & 0.035 &  0.013 \\
    & Acc. $\uparrow$ & 97.01 & 89.27 & 57.25 \\
    \midrule
    \multirow{2}{*}{\textbf{Ours}} & RMSE $\downarrow$ & 0.166 & 0.093 & 0.041 \\
    & Acc. $\uparrow$ & \textbf{97.17} & \textbf{89.35} & \textbf{57.38} \\
    \bottomrule
\end{tabular}
\caption{Comparasion of our method with Remez \cite{zimerman2024converting} and OLS \cite{zheng2022keyword}  for SiLU approximation.}
\label{TabSiLU}
\end{table}
To evaluate the effectiveness of our approach, we compare it with the Remez approximation method used by Zimerman \textit{et al.}~\shortcite{zimerman2024converting} and the least squares method used by Zheng \textit{et al.}~\shortcite{zheng2022keyword} in current privacy-preserving inference works. 
For a fair comparison, we restrict all methods to approximate within the same polynomial degree and range for each dataset. \cref{TabSiLU} presents the results show that our method achieves the highest approximation accuracy across all datasets. For example, on the MNIST dataset, our method achieves an approximation accuracy of $97.17\%$, which is $0.41\%$ and $0.16\%$ higher than the Remez and the least squares method, respectively. 
Because our method prioritizes fitting densely distributed data regions while downplaying sparse edge regions, our method achieves better approximation in regions where the data is more concentrated. As shown in \cref{figSilu}, our method results in a polynomial that more closely approximates in $[-4, 1]$.

\subsection{B-spline Basis Approximation}

\begin{table}[t]
\centering
\begin{tabular}{c|cc|c}
    \toprule
    $(n_i, g, k)$ & Naive (s) & Ours (s) & Speedup ($\times$) \\
    \midrule
    (64, 3, 2)& 54.009 & \textbf{32.691} & 1.65\\
    
    (128, 5, 3) & 71.760 & \textbf{33.527} & 2.14 \\
    
    (256, 5, 3) & 109.712 & \textbf{33.788} & 3.25 \\
    
    (256, 10, 3) & 187.691 & \textbf{33.852} & 5.54 \\
    
    (256, 10, 5) & 212.572 & \textbf{42.241} & 5.03 \\
    \bottomrule
\end{tabular}
\caption{Comparison of the HE inference latency between our B-spline basis approximation method and the naive implementation without lazy combination optimization.}
\label{TabBspline}
\end{table}

\begin{table}[t]
    \centering
    \begin{tabular}{c|c|ccc}
    \toprule
    Tasks & Dataset & Baseline & MLPs & \textbf{Ours} \\
    \midrule
    \multirow{3}{*}{Image $\uparrow$} & MNIST & 97.53 & 97.20 & \textbf{97.27} \\
    & FMNIST & 89.56 & 87.10 & \textbf{89.35} \\
    & CIFAR-10 & 57.62 & 54.67 & \textbf{57.37} \\
    \midrule
    \multirow{3}{*}{Formula $\downarrow$} & Toy & 0.00158 & 0.14072 & \textbf{0.00157} \\
    & lpmv0 & 0.00460 & 0.01404 & \textbf{0.00461} \\
    & lpmv1 & 0.03478 & 0.07378 & \textbf{0.03489} \\
    \bottomrule
    \end{tabular}
    \caption{Comparison of model performance across different datasets. For image classification, we evaluate model performance using accuracy, while for symbolic functions, we use RMSE against the original formula.}
    \label{TabAcc}
\end{table}

Compared to the naive method without lazy combination optimization, our proposed B-spline basis approximation method offers a significant efficiency improvement.
\cref{TabBspline} demonstrates that our method improves computational efficiency several times compared to the naive method. For example, when $n_i=256$, $g=10$, and $k=5$, our method takes only $42.241$ seconds, which is $5.03 \times$ faster than the naive method. This improvement is because the naive method requires a permutation operation after B-spline basis approximation to rearrange the data. The permutation in HE domain introduces additional costs, including $n_i \cdot (g + k)$ HE ciphertext multiplications and $\sqrt{n_i \cdot (g + k)}$ HE rotations, leading to significant latency overhead. As the input dimension $n_i$, B-spline grid $g$, and B-spline degree $k$ increase, the speedup of our method improves.

Additionally, the results in \cref{TabBspline} show that, as the parameters $n_i$ and $g$ increase, the HE inference latency remains relatively stable with our method. This is because our method is designed to take full advantage of SIMD of CKKS for parallel processing. Therefore, as long as the single ciphertext packing condition $n_i \cdot (g + 2k) \leq \frac{N}{2}$ is satisfied, increases in $n_i$ and $g$ do not significantly affect the HE inference latency. 
However, as parameter $k$ increases, we needs to iteratively compute higher-order B-spline basis functions, increasing the HE inference latency due to the added multiplicative depth. In practical applications, the performance of the B-spline basis approximation can be optimized by carefully selecting appropriate values for $n_i$, $g$, and $k$.

\subsection{Comparison of Model Performance}

To evaluate the practical performance of our proposed privacy-preserving KAN inference scheme, we conducted experiments on various datasets. 
We compared with plaintext KAN as a baseline, as well as with MLP \cite{rumelhart1986learning}, which is a fundamental component in current HE-based privacy-preserving inference methods~\cite{benaissa2021tenseal,lee2022low,ran2023spencnn}.
Apart from image classification tasks, we also conducted experiments on symbolic formulas. We selected the following symbolic functions from \cite{liu2024kan} for our experiments. 
The Toy Formula dataset represents a high-dimensional symbolic function $f(x_1,\cdots,x_{100})=\exp(\frac{1}{100}\sum_{i=1}^{100}\sin^2(\frac{\pi x_i}{2}))$, while the lpmv0 and lpmv1 datasets consist of special functions from \cite{liu2024kan}.


The experimental results show that the accuracy of our privacy-preserving KAN inference is very close to that of the plaintext domain and exceeds that of MLPs, as shown in \cref{TabAcc}. On image benchmark datasets, the accuracy decrease of our scheme is no more than $0.25\%$, and it outperforms MLPs. On the symbolic formula dataset, the RMSE deviation from plaintext is no more than $0.00011$, with a lower RMSE than MLPs, confirming the effectiveness of our HE-based KAN inference scheme.


\begin{figure}[t]
\centering
\includegraphics[width=0.9\columnwidth]{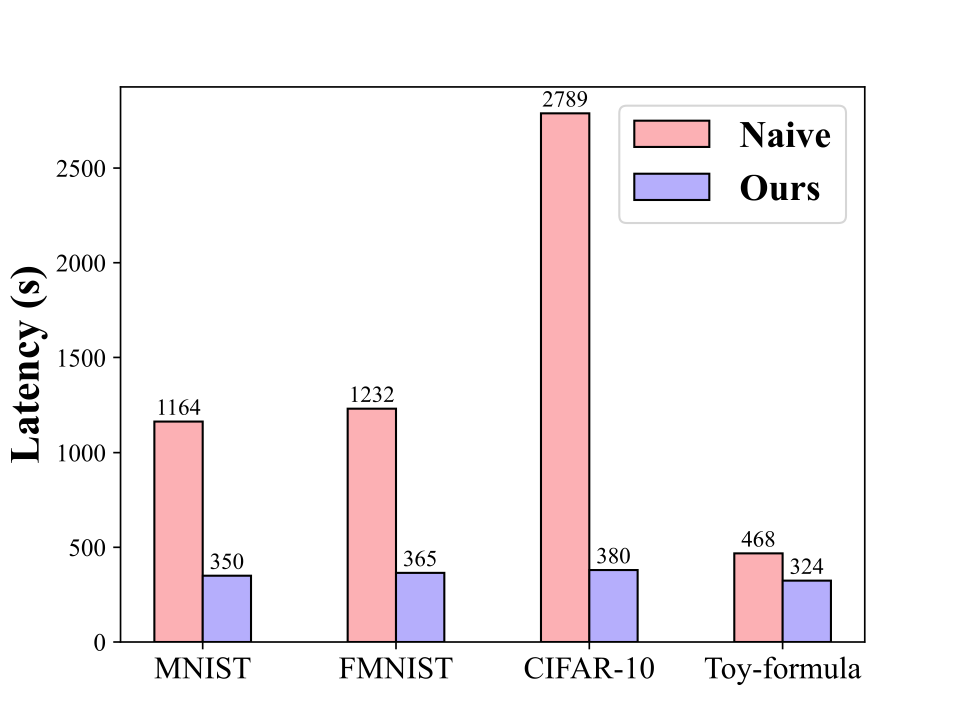} 
\caption{Comparison of inference latency between our method and the naive implementation across different datasets}
\label{figPie}
\end{figure}

\subsection{Inference Latency}
We evaluated the inference latency for a single input across different datasets in a single-threaded environment, comparing our method with the naive implementation, as shown in the figure. For the small dataset Toy-formula, our method achieves an inference latency of 324 seconds, while the naive method has a latency of 468 seconds, resulting in approximately $1.44\times$ speedup. On the larger CIFAR-10 dataset, our inference latency is 380 seconds compared to 2789 seconds for the naive method, yielding over $7\times$ speedup. Additionally, since servers often handle data from multiple users, the amortized processing time for multiple datasets is also important. Due to page limitation, we provide details on the multi-threaded environment's amortized inference latency and the time distribution of various computational components in the Appendix.

\section{Conclusion}
We have proposed an efficient scheme for privacy-preserving KAN inference using HE. We designed a novel task-specific activation function approximation method to closely approximate the SiLU activation function. This approximation enhances the performance of HE networks on real-world tasks without necessitating retraining of the network model.
Additionally, we introduced an efficient method for computing B-spline functions in the HE domain. By leveraging techniques such as repeated packing, lazy composition, and comparison functions, we achieved high-precision B-spline function computations.
Building on these two approximation methods, we developed an HE-based KAN model inference scheme. Our approach delivers about a $7\times$ speedup compared to the naive implementation in the HE domain, while maintaining a minimal accuracy loss of only $0.25\%$ compared to plaintext KANs on the CIFAR-10 dataset.
Given the efficiency constraints of HE, in the future, we plan to explore more effective slot utilization strategies and more compact cryptographic parameters to further enhance efficiency.

\clearpage


\bibliography{ref}

\end{document}